%% file: acl_latex.tex
\pdfoutput=1

\documentclass[11pt]{article}

\usepackage[final]{acl}

\usepackage{times}
\usepackage{latexsym}
\usepackage{enumitem}

\usepackage[T1]{fontenc}

\usepackage[utf8]{inputenc}

\usepackage{microtype}

\usepackage{inconsolata}

\usepackage{graphicx}

\usepackage{amsmath}
\usepackage[most]{tcolorbox}

\input{math_commands.tex}

\usepackage{booktabs}
\usepackage{amsmath}
\usepackage{amssymb}
\usepackage{makecell}
\usepackage{multirow}
\usepackage{xcolor}
\usepackage{pifont}
\usepackage[table]{xcolor}
\usepackage{arydshln}

\newcommand{\okmark}{{\textbf{\color{green}{\ding{51}}}}}
\newcommand{\ngmark}{{\textbf{\color{red}{\ding{55}}}}}

\newcommand{\Manyifeval}{ManyIFEval}
\newcommand{\Stylembpp}{StyleMBPP}
\makeatletter
\def\thanks#1{\protected@xdef\@thanks{\@thanks
        \protect\footnotetext{#1}}}
\makeatother

\title{When Instructions Multiply: Measuring and Estimating LLM Capabilities of Multiple Instructions Following}

\author{
 \textbf{Keno Harada\textsuperscript{1}},
 \textbf{Yudai Yamazaki\textsuperscript{2*}\thanks{* Equal contribution}},
 \textbf{Masachika Taniguchi\textsuperscript{3*}},
 \textbf{Edison Marrese-Taylor\textsuperscript{1}},
\\
 \textbf{Takeshi Kojima\textsuperscript{1}},
 \textbf{Yusuke Iwasawa\textsuperscript{1}},
 \textbf{Yutaka Matsuo\textsuperscript{1}}
\\
 \textsuperscript{1}The University of Tokyo,
 \textsuperscript{2}Kyoto University,
 \textsuperscript{3}University of the Ryukyus
\\
 \small{
 \texttt{{\{keno.harada, emarrese, t.kojima, iwasawa, matsuo\}@weblab.t.u-tokyo.ac.jp}}}\\
 \small{\texttt{yamazaki.yudai.82m@st.kyoto-u.ac.jp, k248443@eve.u-ryukyu.ac.jp}}
 }
\begin{document}
\maketitle
\begin{abstract}
As large language models (LLMs) are increasingly applied to real-world scenarios, it becomes crucial to understand their ability to follow multiple instructions simultaneously. 
To systematically evaluate these capabilities, we introduce two specialized benchmarks for fundamental domains where multiple instructions following is important: Many Instruction-Following Eval (ManyIFEval) for text generation with up to ten instructions, and Style-aware Mostly Basic Programming Problems (StyleMBPP) for code generation with up to six instructions.
Our experiments with the created benchmarks across ten LLMs reveal that performance consistently degrades as the number of instructions increases. 
Furthermore, given the fact that evaluating all the possible combinations of multiple instructions is computationally impractical in actual use cases, we developed three types of regression models that can estimate performance on both unseen instruction combinations and different numbers of instructions which are not used during training.
We demonstrate that a logistic regression model using instruction count as an explanatory variable can predict performance of following multiple instructions with approximately 10\% error, even for unseen instruction combinations. 
We show that relatively modest sample sizes (500 for ManyIFEval and 300 for StyleMBPP) are sufficient for performance estimation, enabling efficient evaluation of LLMs under various instruction combinations.
\footnote{ManyIFEval/StyleMBPP data and evaluation codes are available at \url{https://github.com/kenoharada/Multiple-Instructions-Following}}
\end{abstract}

\section{Introduction}
Large Language Models (LLMs) have demonstrated remarkable capabilities across various tasks~\citep{gpt3, openai2024gpt4technicalreport, anil2023palm2technicalreport, dubey2024llama3, gemma2, claude35, gemini15}. However, their practical utility depends on their ability to simultaneously follow multiple instructions. Simple task descriptions such as ``Write a piece of code'' or ``Summarize this text'' typically produce outputs with limited practical value. Adding specific instructions transforms these basic prompts into practically useful tools - for instance, code generation with style guide compliance enables effective team collaboration, while text summarization with formatting requirements (e.g., bullet points, length limits) ensures human-readable outputs. These kinds of instructions are ubiquitous in real-world applications, and models must be able to follow multiple instructions to be useful in practice.

While some existing benchmarks have made valuable contributions to evaluate multiple-instructions-following capabilities for real-world applications, there are several limitations towards accurate and fair evaluation. 
For instance, IFEval~\citep{zhou2023instruction} and ComplexBench~\citep{wen2024complexbench} assign a different task description across different number of instructions, which prevents from isolating the effect of instruction counts as a controlled variable.
FollowBench~\citep{jiang-etal-2024-followbench} utilized LLM-as-a-judge for measuring multiple-instructions-following performance, which prevents from reliable and stable verification compared to rule-based and programmable one.
Without controlled experimental design, it becomes difficult to systematically and precisely analyze how the number of instructions impacts model performance.

\begin{figure}[t]
    \centering
    \includegraphics[width=1.0\columnwidth]{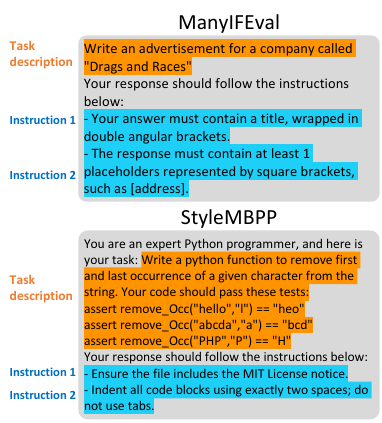}
    \caption{Examples from our proposed \Manyifeval{} (top) and \Stylembpp{} (bottom) benchmarks. Task description defines the primary objective (e.g., text generation, code creation), while instructions specify additional constraints (e.g., formatting rules, style guidelines) that must be simultaneously followed. Our controlled experimental design, where the core task description is kept consistent while varying the number of instructions, allows us to systematically investigate the impact of instruction counts on LLM performance.}
    \label{fig:samples}
\end{figure}

\begin{figure*}[t]
\begin{center}
\includegraphics[width=\textwidth]{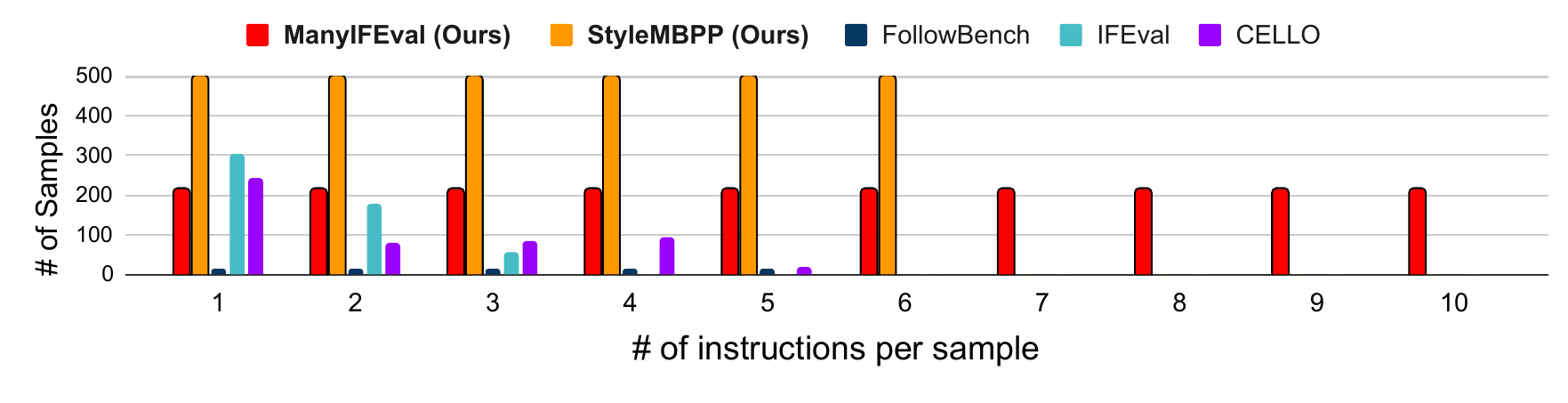}
\end{center}
\vspace{-8mm}
\caption{Comparison of sample size distribution per instruction count between our proposed ManyIFEval and StyleMBPP benchmarks and existing benchmarks (FollowBench, IFEval, and CELLO). Our benchmarks ensure balanced and sufficient sample sizes for each instruction count, enabling a robust analysis of the performance trends as the number of instructions increases.}
\label{fig:histogram}
\end{figure*}

Our work focuses on measuring the simultaneous multiple-instructions-following performance, which we consider crucial in real-world applications. 
Specifically, we introduce two specialized benchmarks: \Manyifeval{} for text generation (up to 10 instructions) extended from IFEval~\citep{zhou2023instruction} and \Stylembpp{} for code generation (up to 6 instructions) extended from MBPP~\citep{mbpp}. 
Our benchmarks maintain consistent task descriptions while varying the number of instructions, allowing us to isolate and analyze the effect of instruction counts on model performance (\autoref{fig:samples} and \autoref{tab:benchmark_comparison}).
Our benchmarks also can objectively measure multiple-instructions-following
ability of LLMs with a programmatically verifiable approach (\autoref{tab:benchmark_comparison}).
In addition, our benchmarks have balanced and sufficient sample sizes for each instruction count (\autoref{fig:histogram}).

Based on the created benchmarks, we conducted a broad range of experiments to evaluate the multiple-instructions-following performance with a wide variety of LLMs. 
Specifically, our experiments used ten state-of-the-art LLMs including API based models (GPT-4o, Claude3.5 Sonnet, Gemini 1.5 Pro, o3-mini) and open models (Llama3.1-8B, Gemma2-9B, Gemma2-2B, Qwen-2.5 72B, DeepSeek-V3, DeepSeek-R1), revealing that performance consistently and drastically degrades as the number of instructions increases.

Furthermore, given the fact that evaluating all the possible combinations of multiple instructions is computationally impractical in actual use cases, we developed regression models that can estimate performance on both unseen instruction combinations and different numbers of instructions which are not used during training.
Specifically, we explore three modeling approaches: naive estimators, beta-binomial, and logistic regression. Through this estimation, we demonstrate that a logistic regression model using instruction count as an explanatory variable can predict performance with approximately 10\% error, even for unseen instruction combinations. 
The approach generalizes effectively to unseen numbers of instructions, achieving a mean absolute error of 0.03 ± 0.04 when predicting performance on 10 instructions using training data from up to 9 instructions.
Our experiments reveal that relatively modest sample sizes are sufficient for accurate performance estimation - 500 samples for ManyIFEval and 300 for StyleMBPP. 
This suggests that our benchmarks can provide reliable performance estimates for LLMs under various instruction combinations, enabling efficient evaluation and guiding future model development.

\section{Related Work}

\begin{table*}[t]
\centering
\scalebox{0.69}{
\begin{tabular}{lcccccccc}
\toprule
\textbf{Benchmark} & \textbf{\#Samples} & 
\textbf{
\begin{tabular}{c}
\#Instructions\\ per sample (max)
\end{tabular}
} & 
\textbf{
\begin{tabular}{c}
Same task description\\ with various \\\# of instructions 
\end{tabular}
} &
\textbf{
\begin{tabular}{c}
Rule-based verifier\\for\\ all instructions
\end{tabular}
} &
\textbf{
\begin{tabular}{c}
Rule-based verifier\\for\\ task description
\end{tabular}
}
\\
\midrule
FollowBench (Mixed) & 85 & 5 & \okmark \;Yes & \ngmark \;No  & \ngmark \;No\\
IFEval & 541 & 3 & \ngmark \;No & \okmark \;Yes  & \ngmark \;No\\
CELLO & 523 & 6 & \ngmark \;No & \okmark \;Yes  & \ngmark \;No\\
CFBench & 1000 & 14 & \ngmark \;No & \ngmark \;No  & \ngmark \;No\\
ComplexBench & 1150 & 14 & \ngmark \;No & \ngmark \;No  & \ngmark \;No\\
InfoBench & 500 & 15 & \ngmark \;No & \ngmark \;No  & \ngmark \;No\\
CodeIF & 1200 & 21 & \ngmark \;No & \ngmark \;No  & \ngmark \;No\\
\midrule
\Manyifeval{} (Ours) & 2160 & 10 & \okmark \;Yes & \okmark \;Yes  & \ngmark \;No\\
\Stylembpp{} (Ours) & 3000 & 6 & \okmark \;Yes & \okmark \;Yes  & \okmark \; Yes \\
\bottomrule
\end{tabular}
}
\caption{Benchmarks for evaluating multiple-instructions-following ability of LLMs. Our benchmarks (ManyIFEval and StyleMBPP) can conduct fair evaluation with the same task description across different number of instructions. In addition, ours can objectively measure the ability with rule-based, programmatically verifiable approach.}
\label{tab:benchmark_comparison}
\end{table*}

\subsection{Instruction-Following Benchmarks}
Instruction-following capability is a crucial aspect of large language models (LLMs) as it influences how well they can meet human expectations. Many early benchmarks focused on evaluating responses to different types of instructions, using relatively simple prompts~\citep{wang-etal-2022-super, alpaca_eval,zheng2023judging}.

As LLMs are increasingly expected to handle more complex real-world tasks, studies have attempted to evaluate models' instruction-following capabilities from various aspects~\citep{xu2024wizardlm,hayati2024coitest,he2024cello,li2024measuring,wen2024complexbench,jiang-etal-2024-followbench, zhang2024cfbench, qin2025sysbench, he2024multiif, yan2025codeif}. ComplexBench~\citep{wen2024complexbench} categorizes instructions into four types based on their characteristics: ``And'' (requiring multiple conditions to be satisfied simultaneously), ``Chain'' (following steps in a specific order), ``Selection'' (choosing actions based on conditions), and ``Nested'' (combining And, Chain, and Selection in a hierarchical structure). The ``And'' type closely aligns with our focus on satisfying multiple instructions simultaneously. 
Additionally, there are existing benchmarks that assess LLMs' multiple-instructions-following ability, as described in \autoref{tab:benchmark_comparison} and \autoref{app:multi}. Multi-turn benchmarks such as Multi-IF~\citep{he2024multiif} and SysBench~\citep{qin2025sysbench} evaluate whether all instructions are followed throughout the entire conversation. In contrast, our benchmark focuses on a more fundamental single-turn problem setting, which is the basis for the multi-turn instruction following ability.

While existing benchmarks use different terminology to refer to the additional requirements that models must satisfy, with some using "constraint" as in ComplexBench~\citep{wen2024complexbench} and FollowBench~\citep{jiang-etal-2024-followbench}, and others using "instruction" as in IFEval~\citep{zhou2023instruction} and Multi-IF~\citep{he2024multiif}, we adopt the term "instruction" following IFEval's convention throughout this paper.

\subsection{Instruction-Following Performance with Multiple Instructions}
While results of previous benchmarks have suggested that following multiple instructions becomes more challenging as their number increases, existing evaluations face limitations in isolating and measuring this effect. For instance, ComplexBench~\citep{wen2024complexbench} shows lower performance on Chain, Selection, and Nested instruction types which contain more instructions, but this could be attributed to various factors beyond instruction count, such as task complexity. FollowBench's Mixed dataset~\citep{jiang-etal-2024-followbench}, while conceptually similar to our focus, has limited samples (17 per instruction count) and relies on model-based evaluation, leading to inconsistent results where higher instruction levels sometimes outperform lower ones. IFEval~\citep{zhou2023instruction} provides objective programmatic verification but varies both instruction count and task descriptions across samples, making it difficult to attribute performance changes specifically to increased instruction count. 
In summary, existing benchmarks have the following three limitations: (1) unbalanced or insufficient sample sizes across instruction counts, (2) reliance on LLM-based evaluation methods that lack objectivity and reliability~\citep{zhou2023instruction,zeng2024evaluatingif,zheng2023judging,wang-etal-2024-large-language-models-fair}, and (3) inconsistency in task descriptions across varying instruction counts.

Our benchmarks are specifically designed to address these limitations by systematically varying the number of instructions added to each task description, ensuring balanced sample sizes across different instruction counts. In addition, by employing programmatic rule-based verification for each added instruction, we enable objective and reliable assessment of the relationship between instruction count and instruction-following performance.
See \autoref{tab:benchmark_comparison} and \autoref{fig:histogram} for details.

\subsection{Performance Estimation}
As parameters of language models grow, the computational cost of evaluation on benchmarks increases. While some studies attempt to reduce evaluation costs through strategic sampling~\citep{perlitz-etal-2024-efficient,vivek-etal-2024-anchor,polo2024tinybenchmarks} or combining existing benchmarks~\citep{Jinjie2024mixeval}, we focus on modeling and estimating how instruction-following performance degrades as the instruction count increases, using our benchmarks.

\section{Benchmark Creation}
\label{sec:benchmark}
To systematically investigate multiple instructions-following capabilities of LLMs, we introduce two benchmarks: ManyIFEval for general writing task with up to ten instructions and StyleMBPP for code generations with up to six instructions. Both benchmarks are constructed with programmatically verifiable instructions to ensure objective and reliable evaluation. Our benchmarks are specifically created using sets of instructions that are non-conflicting and simultaneously satisfiable. The primary goal is to investigate how performance degrades as the number of instructions increases, keeping the instructions themselves compatible. We ensured this compatibility through careful curation. \Manyifeval{} extends IFEval~\citep{zhou2023instruction}, while \Stylembpp{} builds upon MBPP~\citep{mbpp}.
ManyIFEval and StyleMBPP are under the CC BY 4.0 International license same as original MBPP and IFEval.

\subsection{ManyIFEval}
\Manyifeval{} (Many Instruction-Following Eval) is designed to evaluate text generation tasks under multiple instructions, and is an extension of IFEval~\citep{zhou2023instruction}, which comprises task descriptions such as \textit{"Write a blog post about a trip to Japan."} and a set of objectively verifiable instructions shown in \autoref{fig:samples} (top).

We started with the 25 instructions from the original IFEval dataset. We removed instructions that were inherently difficult to combine or prone to conflict (e.g., specifying the output language, repeating the prompt, generating two distinct responses) or had very low individual success rates (e.g., specifying first word in i-th paragraph). This resulted in a set of 15 instructions. We mapped out potential conflicts between these 15 (e.g., character count limits vs. paragraph count requirements; uppercase vs. lowercase). When generating samples for our benchmark, we algorithmically ensured that only mutually compatible instructions were selected and combined for any given prompt.

Finally, we selected 216 task descriptions from IFEval. For each task description, we generated samples with varying numbers of non-conflicting and simultaneously satisfiable instructions, ranging from 1 to 10. The instructions in ManyIFEval fall into six categories:
(1) keyword requirement for including/excluding specific words, (2) length instructions for controlling text length, (3) formatting requirements like bullet points and placeholders, (4) letter case specifications for uppercase/lowercase usage, (5) text beginning/ending requirements like quotation marks, and (6) punctuation rules for allowing/prohibiting specific marks. \autoref{tab:manyifeval-single} in \autoref{app:each_instruction} shows the success rates of individual instructions. 

To demonstrate that rule-based verification is more reliable than LLM-as-a-Judge, we compared both evaluation methods on ManyIFEval. As shown in Table~\ref{tab:llm_judge_comparison}, LLM-as-a-Judge tends to inflate accuracy scores, particularly as instruction count increases. This inflation makes it difficult to accurately measure the true degradation in multiple-instructions-following capabilities, reinforcing the importance of objective, programmatic verification for reliable benchmark evaluation.

\begin{table}[t]
\centering
\scalebox{0.9}{
\begin{tabular}{ccc}
\toprule
\# Instructions & Rule-based & LLM-as-a-Judge \\
& (Ground Truth) & (GPT-4o) \\
\midrule
5 & 0.574 & 0.815 \\
10 & 0.213 & 0.657 \\
\bottomrule
\end{tabular}
}
\caption{Comparison of Prompt-level Accuracy on ManyIFEval using rule-based verification vs. LLM-as-a-Judge (GPT-4o zero-shot). LLM-as-a-Judge tends to inflate accuracy scores and makes it difficult to accurately measure performance degradation as instruction count increases.}
\label{tab:llm_judge_comparison}
\end{table}

\subsection{StyleMBPP}
\Stylembpp{} (Style-aware Mostly Basic Programming Problems) is designed for evaluating code generation tasks under multiple style-related instructions, as shown in \autoref{fig:samples} (bottom). This benchmark extends MBPP~\citep{mbpp}, a code generation benchmark focused on basic Python programming problems, such as \textit{"Write a Python function to sort the given array by using merge sort."}. While MBPP assesses code correctness based on test cases, \Stylembpp{} adds instructions to evaluate models' adherence to coding style guidelines while maintaining functional correctness. We selected common Python coding style guidelines, primarily focusing on those verifiable using Pylint~\citep{pylint}. We chose non-conflicting instructions and filtered out any with extremely low individual success rates. Finally, we add a new instruction to this original set, where we require the model to incorporate a ``MIT License notice'' to the generated code.

For \Stylembpp{}, we augmented the 500 test problems from MBPP with style instructions. For each MBPP problem, we created samples with instruction counts ranging from 1 to 6, resulting in 3,000 samples. The style instructions require specific code formatting aspects: (1) Generated code files must include the MIT License notice. (2) Code must follow strict indentation rules, using spaces instead of tabs. (3) Functions must include docstrings. (4) Specific conditional comparison operators may be required or prohibited. (5) Lines must not exceed a maximum character length. (6) Variable names must adhere to length restrictions. 
\autoref{tab:stylembpp-single} in \autoref{app:each_instruction} shows the success rate of following individual instructions.

\section{Evaluation}
\subsection{Setup}
To evaluate the multiple-instructions-following capabilities of LLMs, we conducted experiments using both closed and open models. For closed models (accessed via APIs), we evaluated GPT-4o (gpt-4o-2024-05-13)~\citep{gpt4o}, Claude 3.5 Sonnet (claude-3-5-sonnet-20240620)~\citep{claude35}, Gemini 1.5 Pro (gemini-1.5-pro-002)~\citep{gemini15} and o3-mini~\citep{o3-mini}. For open models, we assessed Gemma 2 (gemma-2-9b-it,gemma-2-2b-it)~\citep{gemma2}, Llama 3.1 (Meta-Llama-3.1-8B-Instruct)~\citep{dubey2024llama3}, Qwen2.5-72B (Qwen2.5-72B-Instruct)~\citep{qwen2025qwen25technicalreport}, DeepSeek-V3 (DeepSeek-V3-0324)~\citep{deepseekv3} and DeepSeek-R1~\citep{deepseekr1}.  All models were evaluated using zero-shot prompting presenting the task description along with varying numbers of instructions. See \autoref{app:setup} for the detailed evaluation setup. We report results of Qwen2.5-72B, DeepSeek-V3, DeepSeek-R1 and o3-mini in \autoref{app:detailed_results}.

\subsection{Evaluation Metrics}
To evaluate instruction-following performance, we adopted metrics based on IFEval~\citep{zhou2023instruction} and FollowBench~\citep{jiang-etal-2024-followbench}.
\paragraph{Prompt-level Accuracy}
    Prompt-level Accuracy is the the success rate of following all given instructions simultaneously for a particular prompt (\autoref{eq:prompt_level_accuracy}). This assesses the model's capability to handle multiple instructions at once. This is Hard Satisfaction Rate (HSR) in FollowBench. Prompt-level Accuracy is defined as follows:
\vspace{-3mm}
\begin{equation}
    \text{Prompt-level Accuracy (n)} = \frac{1}{m} \sum_{i=1}^{m} \prod_{j=1}^{n} s_{i}^{j},
    \label{eq:prompt_level_accuracy}
\end{equation}
where \( m \) represents the number of prompts and \( n \) represents the number of instructions per prompt, \( s^{j}_{i} \) represents a binary metric if the target instruction \( j \in n \) in a task \( i \in m \)  is successfully followed.
Therefore, \( \prod_{j=1}^{n} s_{i}^{j} = 1 \) if all instructions for prompt \( i \) are satisfied, and 0 otherwise. In other words, this metric computes whether all the instructions in a prompt have been followed.

\paragraph{Instruction-level Accuracy (Inst-level accuracy)}
Instruction-level Accuracy is the success rate of following individual instructions in its response (\autoref{eq:inst_level_accuracy}). This metric assesses the model's ability to adhere to each instruction separately. This is Soft Satisfaction Rate (SSR) in FollowBench.
This accuracy is defined as follows:
\vspace{-2mm}
\begin{equation}
    \text{Inst-level Accuracy (n)} = \frac{1}{mn} \sum_{i=1}^{m} \sum_{j=1}^{n} s_{i}^{j},
    \label{eq:inst_level_accuracy}
\vspace{-1mm}
\end{equation}
where \( s_{i}^{j} = 1 \) if the \( j \)-th instruction of the \( i \)-th task is satisfied, and \( s_{i}^{j} = 0 \) otherwise.

\subsection{Result}
\begin{figure}[t]
   \centering
   \includegraphics[width=1.0\columnwidth]{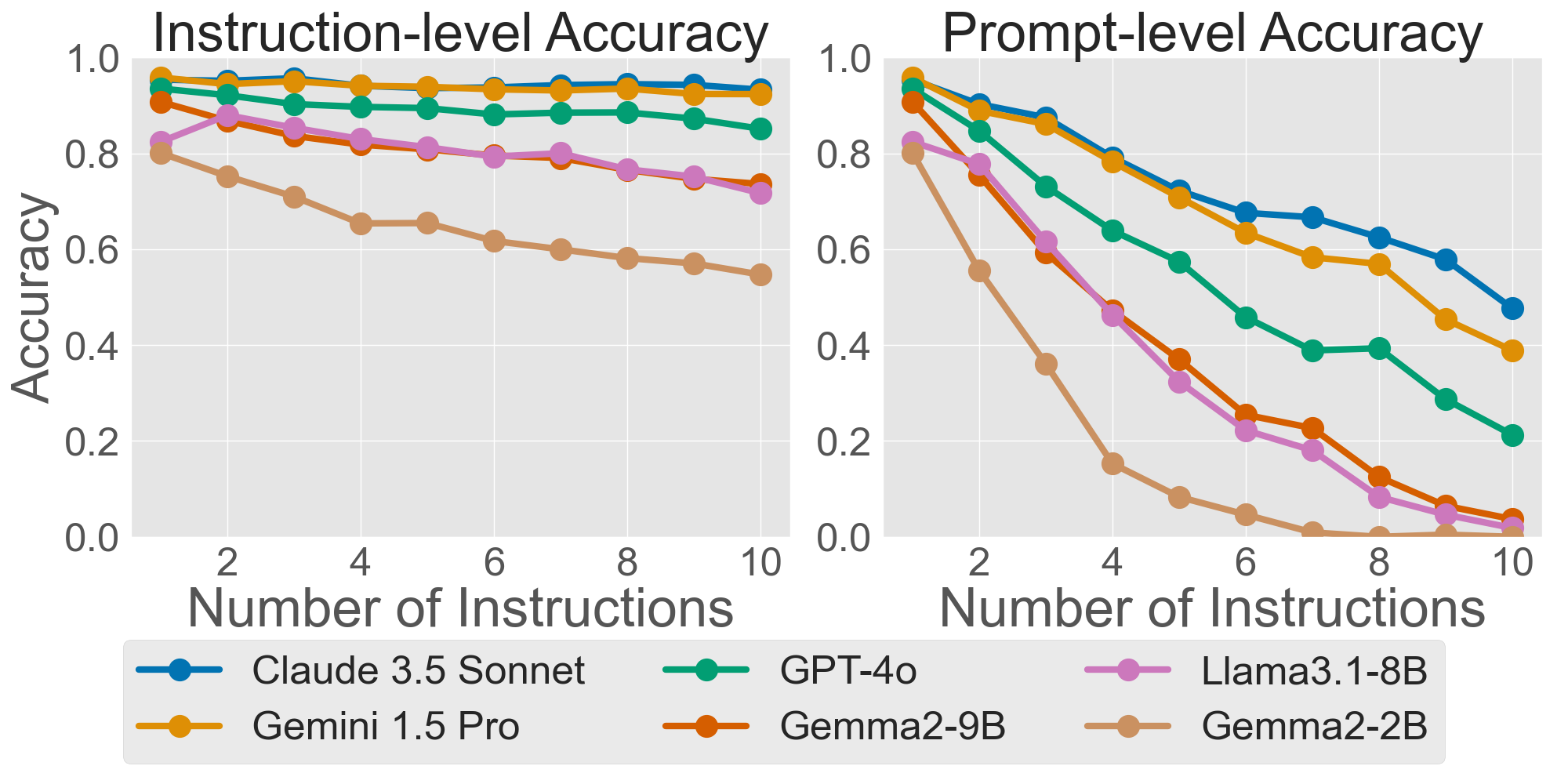}
   \caption{Evaluation results on ManyIFEval. The left panel shows Instruction-level Accuracy (the average success rate of following individual instructions), and the right panel shows Prompt-level Accuracy (the success rate of satisfying all instructions in a prompt simultaneously). Prompt-level Accuracy  consistently shows a degrading trend as the number of instructions increases, while Instruction-level Accuracy remains relatively.}
   \label{fig:results_manyifeval}
\end{figure}

\paragraph{ManyIFEval}
\autoref{fig:results_manyifeval} shows the Instruction-level Accuracy (left panel) and Prompt-level Accuracy (right panel) across different instruction counts for ManyIFEval. Results consistently demonstrate performance degradation across all evaluated models as the number of instructions increases, for both metrics. While models exhibit relatively high accuracy in following individual instructions in isolation, their ability to satisfy all instructions simultaneously diminishes significantly as the instruction count rises. Notably, the relative performance ranking among models remains consistent between Instruction-level and Prompt-level accuracy metrics - models that perform better at following individual instructions also demonstrate superior performance when required to follow multiple instructions simultaneously. This indicates that the presence of multiple instructions introduces complexity for LLMs, even when the individual instructions are intrinsically simple. \autoref{fig:each_instruction_manyifeval} in \autoref{app:each_instruction} shows detailed results of following individual instructions. Through comparison with experimental results from IFEval and FollowBench, we demonstrated that our benchmark provides more stable performance measurement across a higher number of instructions (see \autoref{fig:benchmark_result_comparison} in \autoref{app:comparison}).

\begin{figure}[t]
   \centering
   \includegraphics[width=1.0\columnwidth]{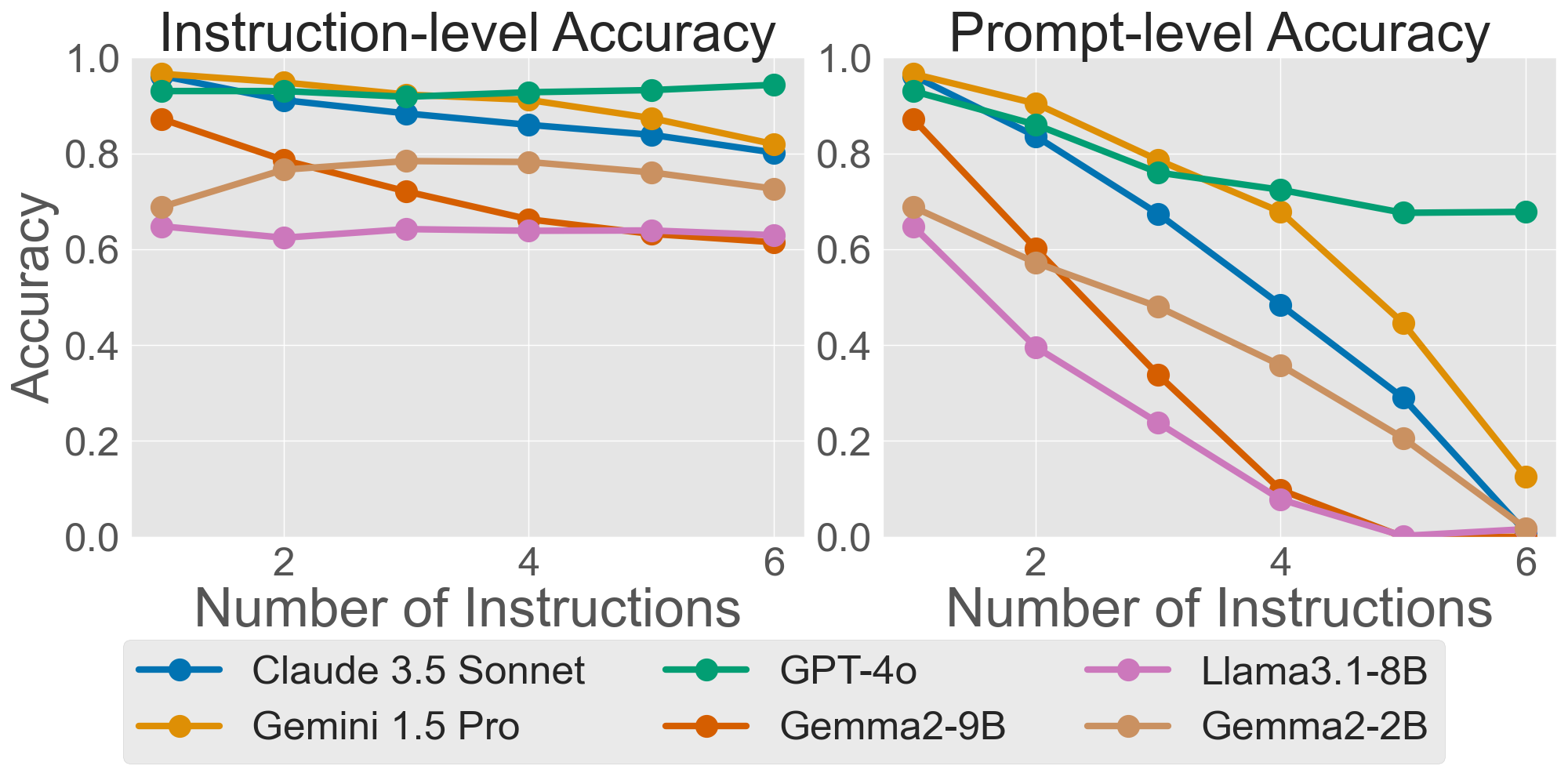}
   \caption{Evaluation results on StyleMBPP. Similar to ManyIFEval, the Prompt-level Accuracy shows consistent degradation trend as the number of instructions increases.}
   \label{fig:results_stylembpp}
\end{figure}
\begin{figure}[t]
   \centering
   \includegraphics[width=1.0\columnwidth]{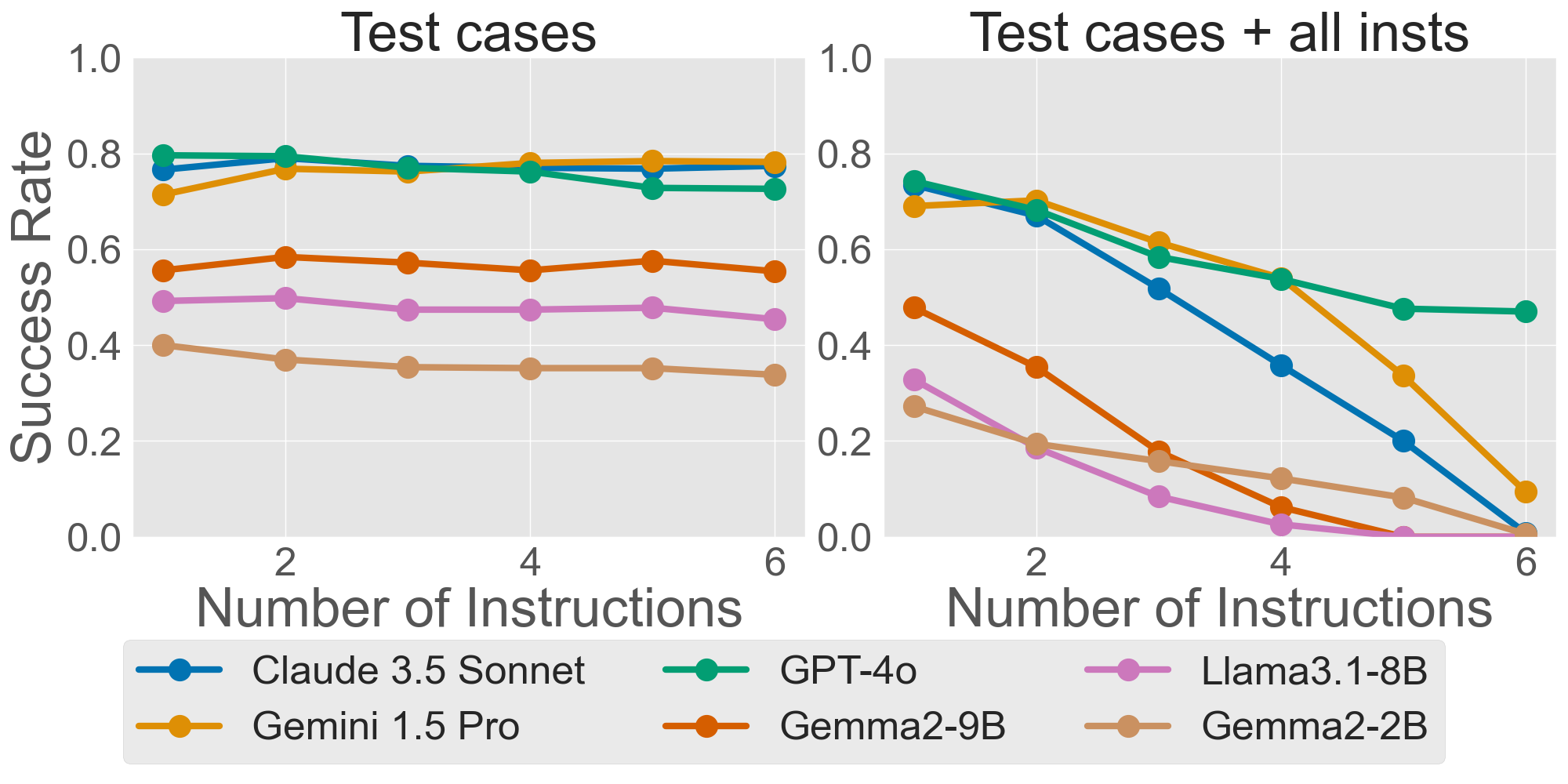}
   \caption{Evaluation results of code generation success rate on StyleMBPP. Left panel: success rate of passing all test cases for the task description, which remains relatively stable even as more instructions are added. Right panel: success rate of passing all test cases and satisfying all instructions simultaneously ("Test cases + all insts"), which shows a significant performance drop as instruction count increases.}
   \label{fig:results_stylembpp_task}
\end{figure}

\paragraph{\Stylembpp{}}
Secondly, we evaluated performance on \Stylembpp{}, which requires models to follow multiple style guide instructions while generating Python code to pass test cases. Result in \autoref{fig:results_stylembpp} shows similar degradation patterns as in \Manyifeval{}. While models exhibit relatively high accuracy in following individual instructions in isolation, their performance of following all instructions degrade as the instruction count rises.

Compared to GPT-4o, both Gemini 1.5 Pro and Claude 3.5 show much lower Prompt-level Accuracy when given six instructions simultaneously. This performance gap is attributed to certain instructions becoming significantly more challenging for these models as the instruction count increases. Specifically, while both models can follow the ``Characters per line'' instruction with high success rates (99\% for Gemini 1.5 Pro and 97\% for Claude 3.5) when it is presented in isolation, their performance drops dramatically to 20\% and 2\% respectively when this instruction is combined with five other instructions. For open models like Gemma 2 and Llama 3.1, we observe that their success rate for the "Indentation" instruction is extremely low, making it difficult for these models to follow all instructions successfully when this instruction is included in the prompt. \autoref{fig:each_instruction_stylembpp} in \autoref{app:each_instruction} shows detailed results of following individual instructions.

As shown in \autoref{fig:results_stylembpp_task}, the functional correctness of the generated code measured by passing all test cases remains relatively stable even as more style instructions are added (left panel). However, for more practical cases where a base task description must be executed while following multiple instructions simultaneously, we observe performance degradation as the number of instructions increases (right panel). This indicates that while models can maintain their core programming capabilities, their ability to simultaneously satisfy multiple style guide instructions becomes increasingly challenging with more instructions.

\paragraph{Improvements by Reasoning}
As shown in Tables~\ref{tab:manyifeval_prompt_level}, \ref{tab:manyifeval_instruction_level}, \ref{tab:stylembpp_prompt_level}, \ref{tab:stylembpp_instruction_level} of \autoref{app:detailed_results}, we observed that reasoning models outperform non-reasoning models (DeepSeek-V3 vs DeepSeek-R1, GPT-4o vs o3-mini). We observed that setting "reasoning effort" in o3-mini to "high" yielded the best results. In reasoning traces, DeepSeek-R1 explicitly checks each given instruction one by one to formulate a plan of approach as shown in \autoref{tab:example_reasoning_trace}. This suggests that reasoning helps models better understand and adhere to multiple instructions.

\paragraph{Comparison with General Benchmarks}
While models show comparable performance on standard benchmarks such as MMLU~\citep{hendrycks2021mmlu}, our benchmarks reveal substantial performance gaps on the instruction-following capabilities. For instance, DeepSeek-V3, GPT-4o, and Qwen2.5-72B achieve similar scores on standard benchmarks but there are performance gaps measured by Prompt-level Accuracy on ManyIFEval (n=10) and on StyleMBPP (n=6). See \autoref{app:performance_gap} for detailed comparison.

\section{Performance Prediction}
As parameters of models grow, their evaluation becomes increasingly computationally expensive. In multiple-instruction settings, evaluating all of them would require substantial computational resources because the number of possible instruction combinations are massive. To address this challenge, we develop models capable of estimating performance on both unseen instruction combinations and different numbers of instructions. We explore three modeling approaches: \textbf{Naive Estimators}, \textbf{Beta-Binomial}, and \textbf{Logistic Regression}, each offering different perspectives on modeling multiple-instructions-following behavior.

\subsection{Models}
\paragraph{Naive Estimators}
We model the success or failure of following a single instruction as a Bernoulli trial with probability \(p \in [0, 1]\). For multiple instructions presented together, the simplest approach assumes independent Bernoulli trials. The probability of successfully following all \(n\) instructions is the product of their individual success probabilities, $\prod_{i=1}^{n} p_i$, where \(p_i\) represents instruction \(i\)'s success probability. We explore two variants of this estimator. In the first variant, which we refer to as \textbf{Product(Each, n=1)}, we use the empirical success rates of following individual instructions when each instruction is presented in isolation. In the second variant, referred to as \textbf{Product(Each, n=n)}, we use the empirical success rates of following individual instructions when the instruction is presented together with \(n-1\) other instructions.

\paragraph{Beta-Binomial}
Similar to Naive Estimators, the Beta-Binomial approach models each instruction’s success as a Bernoulli trial with probability \(p\), but treats \(p\) itself as a random variable drawn from a Beta distribution \(p\sim\mathrm{Beta}(\alpha,\beta)\), and estimates \(\alpha\), \(\beta\) via maximum likelihood from the training data.

\paragraph{Logistic Regression}
We train logistic regression models that predict the probability of following all instructions successfully, based on features such as the number of instructions and instruction identifiers. We investigate several configurations of the model. One configuration, denoted as \textbf{Logistic (w/ n)}, includes only the instruction count as a feature. Another, referred to as \textbf{Logistic (w/ n) (trained n$\leq$k)}, is trained on data containing at most \(k\) instructions, also using instruction count as a feature. A third configuration, \textbf{Logistic (w/ n, IDs)}, incorporates both the instruction count and instruction identifiers, unique numerical labels assigned to each distinct type of instruction, as features.

Beta-Binomial and Logistic (w/ n) models do not account for each instruction's difficulty, while Product and Logistic (w/ n, IDs) incorporate this information through instruction identifiers.

\vspace{-1mm}
\subsection{Training and Evaluation Setup}
\vspace{-1mm}
For all methods, we evaluate performance using the predicted probability of successfully following all instructions, compared to the actual success rate measured by Prompt-level Accuracy on the test set. This allows us to assess how well each approach generalizes to both unseen instruction combinations and different number of instructions.
For both ManyIFEval and StyleMBPP benchmarks, we split the datasets into train and test sets to fit the models on the train split and evaluate their performance on the test split. When splitting the data, we focused on task descriptions and instruction combinations. Specifically, for ManyIFEval, we created sets of instruction combinations with 5 instructions, while for StyleMBPP, we created sets with 3 instructions. We then divided these combination sets between train and test splits ensuring no overlap in combinations. The splits were created such that there were no shared task descriptions between train and test sets, and no shared instruction combinations for the 5-instruction cases in ManyIFEval and 3-instruction cases in StyleMBPP. This resulted in 1,070 training samples and 1,090 test samples for ManyIFEval, and 1,458 training samples and 1,542 test samples for StyleMBPP. 

We estimate parameters for the Beta-Binomial model using NumPy and SciPy~\citep{harris2020array,2020SciPy-NMeth}, while the logistic regression model is trained with statsmodels~\citep{seabold2010statsmodels}. We used GPT-4o, Claude 3.5 Sonnet,  Gemini 1.5 Pro,  Gemma 2 (2B, 9B) and Llama 3.1 for experiments.

\subsection{Results}
\begin{figure}[t]
    \centering
    \includegraphics[width=1.0\columnwidth]{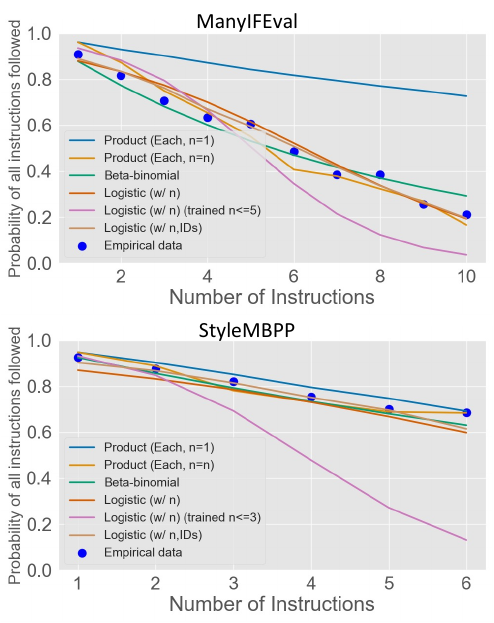}
    \caption{Comparison of empirical Prompt-level Accuracy data (points) and predictions from various estimation models (lines) on ManyIFEval (top) and StyleMBPP (bottom) for GPT-4o. Simple models such as Beta-binomial, Logistic regression model using instruction count as a feature capture the performance degradation trend as the number of instructions increases.}
    \vspace{-2mm}
    \label{fig:estimation_manyifeval_and_stylembpp}
\end{figure}

\begin{figure}[t]
    \centering
    \includegraphics[width=1.0\columnwidth]{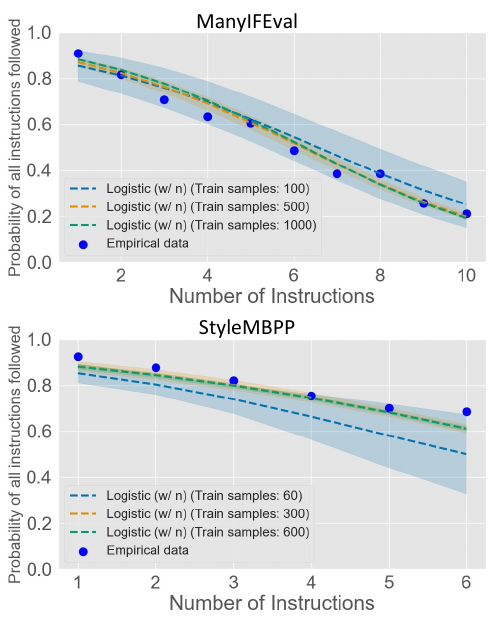}
    \caption{Train sample size variation and estimation error of ManyIFEval (top) and StyleMBPP (bottom) tested on GPT-4o. ManyIFEval requires 500 samples (50 task descriptions for each \# of instructions) and StyleMBPP requires 300 samples (50 task descriptions for each \# of instructions) to achieve stable estimates.}
    \label{fig:size_variation_manyifeval_stylembpp}
    \vspace{-2mm}
\end{figure}

\begin{table*}[t]
  \centering
  \scalebox{0.9}{
  \begin{tabular}{lcccccc}
    \toprule
    \multirow{2}{*}{Method} 
      & \multicolumn{3}{c}{ManyIFEval} 
      & \multicolumn{3}{c}{StyleMBPP} \\
    \cmidrule(lr){2-4} \cmidrule(lr){5-7}
      & n=5 & n=10 & Corr (r) 
      & n=3 & n=6 & Corr (r) \\
    \midrule
    Product (Each, n=1)     & $0.21 \pm 0.07$ & $0.34 \pm 0.13$ & 0.904 & $0.15 \pm 0.09$ & $0.31 \pm 0.32$ & 0.728 \\
    Product (Each, n=n)     & $0.04 \pm 0.03$ & $0.02 \pm 0.02$ & 0.994 & $0.05 \pm 0.01$ & $0.01 \pm 0.01$ & 0.996 \\
    Beta-binomial           & $0.05 \pm 0.05$ & $0.06 \pm 0.03$ & 0.974 & $0.05 \pm 0.02$ & $0.19 \pm 0.13$ & 0.922 \\
    Logistic (w/ n)         & $0.04 \pm 0.03$ & $0.02 \pm 0.03$ & 0.993 & $0.06 \pm 0.05$ & $0.05 \pm 0.03$ & 0.980 \\
    Logistic (w/ n, IDs)     & $0.03 \pm 0.04$ & $0.02 \pm 0.03$ & 0.994 & $0.05 \pm 0.05$ & $0.03 \pm 0.02$ & 0.988 \\
    \bottomrule
  \end{tabular}
  }
  \caption{
  Mean absolute error ± standard deviation and Pearson correlation (r) of Prompt-level Accuracy predictions by various performance estimation models.
  }
  \label{tab:estimation_error}
  \vspace{-3mm}
\end{table*}

\autoref{fig:estimation_manyifeval_and_stylembpp} illustrates the empirically-observed Prompt-level accuracy and the values predicted by models on ManyIFEval and StyleMBPP. As shown in \autoref{tab:estimation_error}, despite its simplicity, the Logistic (w/ n) model achieves mean absolute errors consistently within 0.1 across both ManyIFEval and StyleMBPP benchmarks, demonstrating effective generalization to unseen instruction combinations.

To assess the relationship between sample size and estimation error, we conducted an analysis varying the size of the training set, as shown in \autoref{fig:size_variation_manyifeval_stylembpp}. For each training sample size, we run five iterations changing random seeds. Our results demonstrate that relatively modest sample sizes are sufficient for reliable performance estimation. Specifically, ManyIFEval requires approximately 500 samples (50 task descriptions for each number of instructions) to achieve stable estimates, while StyleMBPP needs around 300 samples (50 task descriptions for each number of instructions). Beyond these thresholds, increasing the sample size yields diminishing returns in terms of estimation accuracy. This finding has practical implications for future benchmark development and evaluation, suggesting that comprehensive multiple-instructions-following assessment can be achieved with manageable dataset size that saves computational resources.

For unseen number of instructions, we evaluated the Logistic (w/ n) model on ManyIFEval. 
\autoref{tab:manyifeval_unseen_number} shows that the model exhibits a consistent decrease in estimation error as the number of instructions used for training increases. Same trend is observed for StyleMBPP (See~\autoref{tab:stylembpp_unseen_number} in~\autoref{app:detailed_results}).

\begin{table}
\centering
\scalebox{0.9}{
\begin{tabular}{lll}
\toprule
Logistic (w/ n) & Abs Err (n=10) & Corr (r)\\
\midrule
trained n$\leq$5& $0.15 \pm 0.16$ & 0.944\\
trained n$\leq$7& $0.09 \pm 0.09$ & 0.989\\
trained n$\leq$9& $0.03 \pm 0.04$ & 0.988\\
\bottomrule
\end{tabular}
}
\caption{Result of estimation for unseen number of instructions. Absolute error between predictions and empirical result of successfully following all instructions on unseen number of instructions aggregated over six LLMs on ManyIFEval.}
\vspace{-2mm}
\label{tab:manyifeval_unseen_number}
\end{table}

\subsection{Discussion}
\vspace{-1mm}
Although our study has systematically analyzed multiple-instructions-following ability of LLMs, several important questions remain for future work. First, whether similar relationships between instruction count and performance hold for more complex instruction types not covered in our benchmarks, such as semantic instructions, conditional logic, or multi-step procedures. Second, further investigation is needed into the mechanisms behind the performance degradation observed with increasing instruction count. Based on previous works~\citep{venkateswaran2025spotlightinstructions, heo2025do, stolfo2025improving} which suggested a relationship between instruction following and the activation values of specific neurons or attention scores, understanding these failure modes through analysis of attention patterns and internal model representations could inform the development of models more robust to multiple simultaneous instructions. 

\section{Conclusion}
\vspace{-1mm}
We introduced two benchmarks, ManyIFEval and StyleMBPP, to evaluate models' ability to follow multiple instructions simultaneously.
Our experiments have demonstrated that performance degrades as the number of instructions increases, with the degradation pattern varying across models and instruction types. 
We also showed that this degradation can be modeled using simple approaches like logistic regression, enabling accurate prediction of model performance on unseen instruction combinations with relatively small sample sizes. 
These findings provide insights for understanding and improving models' ability to handle multiple instructions simultaneously, an essential capability for real-world applications.

\section*{Limitations}
While our benchmarks cover a range of instruction types, they are still limited in scope. We focused on relatively simple instructions that can be objectively evaluated, such as keyword inclusion, character counts, and formatting rules. More complex instruction types involving semantic understanding, conditional logic (e.g., if/then rules), or multi-step procedures (e.g., first summarize then analyze) were not included. Second, our analysis of performance degradation mechanisms is primarily empirical; we observe the patterns but cannot fully explain the underlying causes. A deeper investigation of model internals, such as analyzing attention scores between instruction tokens and output tokens, studying activation patterns in different model layers, and examining how token representations evolve during the generation process, would be valuable for understanding why certain models struggle more with multiple instructions. 

\section*{Acknowledgments}
We used ChatGPT for translation and paraphrasing of our original text, and GitHub Copilot to help draft plotting scripts for our experimental visualizations

\bibliography{custom}
\appendix
\section{Relationship between Multi-turn Benchmarks and Our Benchmarks}
\label{app:multi}
Multiple instructions following is a practically important capability, and in recent years, benchmarks have been proposed in multi-turn settings to evaluate whether all instructions are followed throughout the entire conversation such as Multi-IF~\citep{he2024multiif} and SysBench~\citep{qin2025sysbench}. 
In contrast, our benchmark focuses on a more fundamental single-turn problem setting, which is the basis for the multi-turn instruction following ability.

\begin{figure}[t]
   \centering
   \includegraphics[width=1.0\columnwidth]{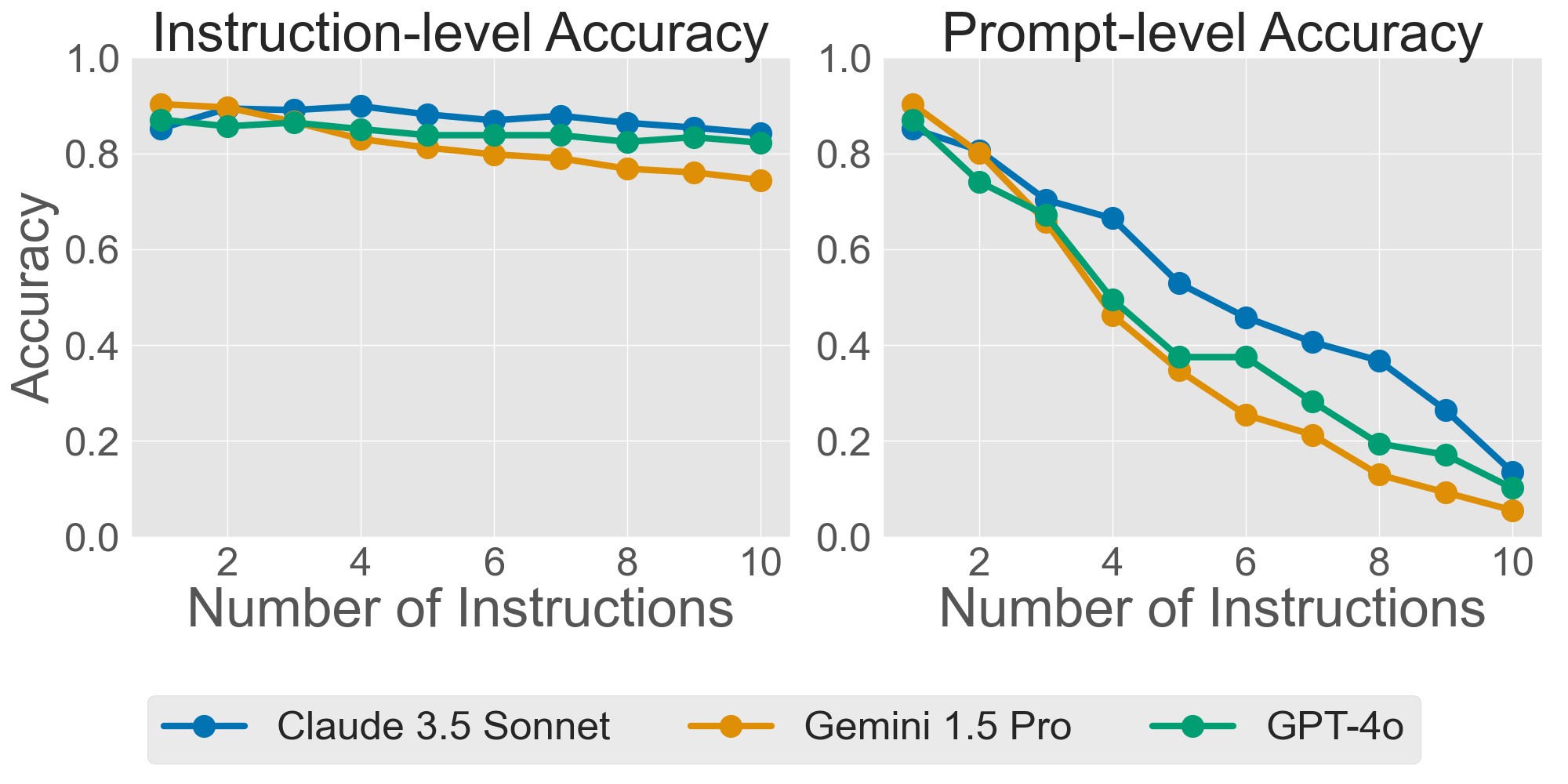}
   \caption{Evaluation results on ManyIFEval in multi-turn settings. Similar to the single-turn setting, we observed that performance in following instructions declines as the number of instructions increases.}
   \label{fig:results_manyifeval_multi}
\end{figure}

\begin{figure}[t]
   \centering
   \includegraphics[width=1.0\columnwidth]{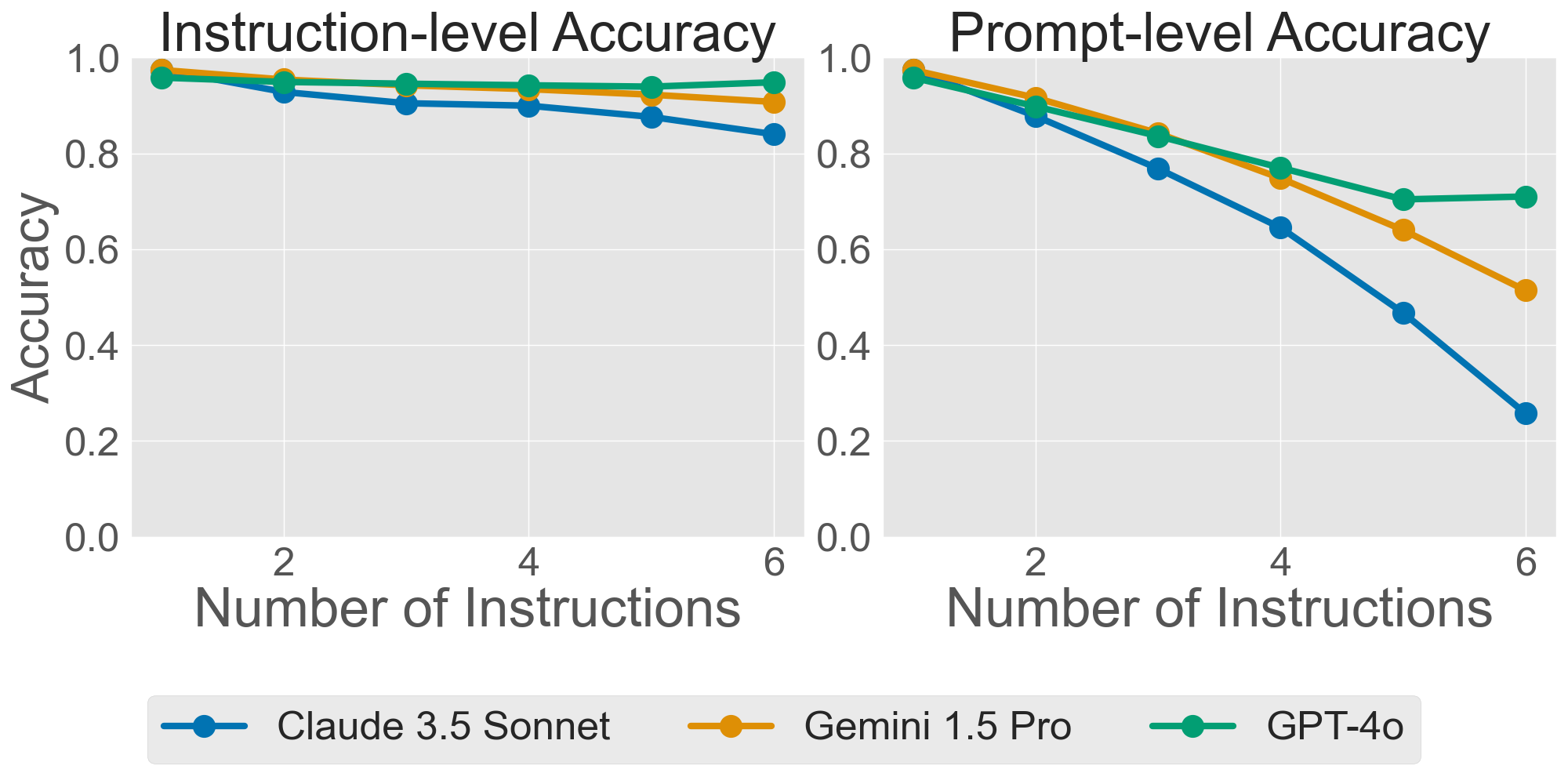}
   \caption{Evaluation results on StyleMBPP in multi-turn settings. Similar to the single-turn setting, we observed that performance in following instructions declines as the number of instructions increases.}
   \label{fig:results_stylembpp_multi}
\end{figure}

Our benchmarks can be easily extended to the problem setting where instructions are added turn by turn, as seen in Multi-IF and SysBench. The results are presented in \autoref{fig:results_manyifeval_multi} and \autoref{fig:results_stylembpp_multi}. Similar to the single-turn setting, we observed that performance in following instructions declines as the number of instructions increases.

\section{Reliable Performance Measurement with Increasing Instruction Counts using ManyIFEval}
\label{app:comparison}
Evaluating the capability of Large Language Models (LLMs) to follow an increasing number of simultaneous instructions requires benchmarks that ensure both reliable verification and sufficient data. Previous benchmarks, such as IFEval and FollowBench, while valuable, have sometimes faced limitations due to smaller sample sizes for certain instruction configurations or reliance on evaluation methods that can introduce variability.

Our benchmarks are specifically designed to address these challenges by combining (1) programmatic, rule-based verification for objective assessment of instruction following, and (2) a substantial and balanced number of samples for each count of instructions. This design ensures that performance changes can be confidently attributed to the model's ability to handle an increasing instruction load, rather than artifacts of small sample sizes or subjective evaluation.

\begin{figure}[t]
    \centering
    \includegraphics[width=0.8\linewidth]{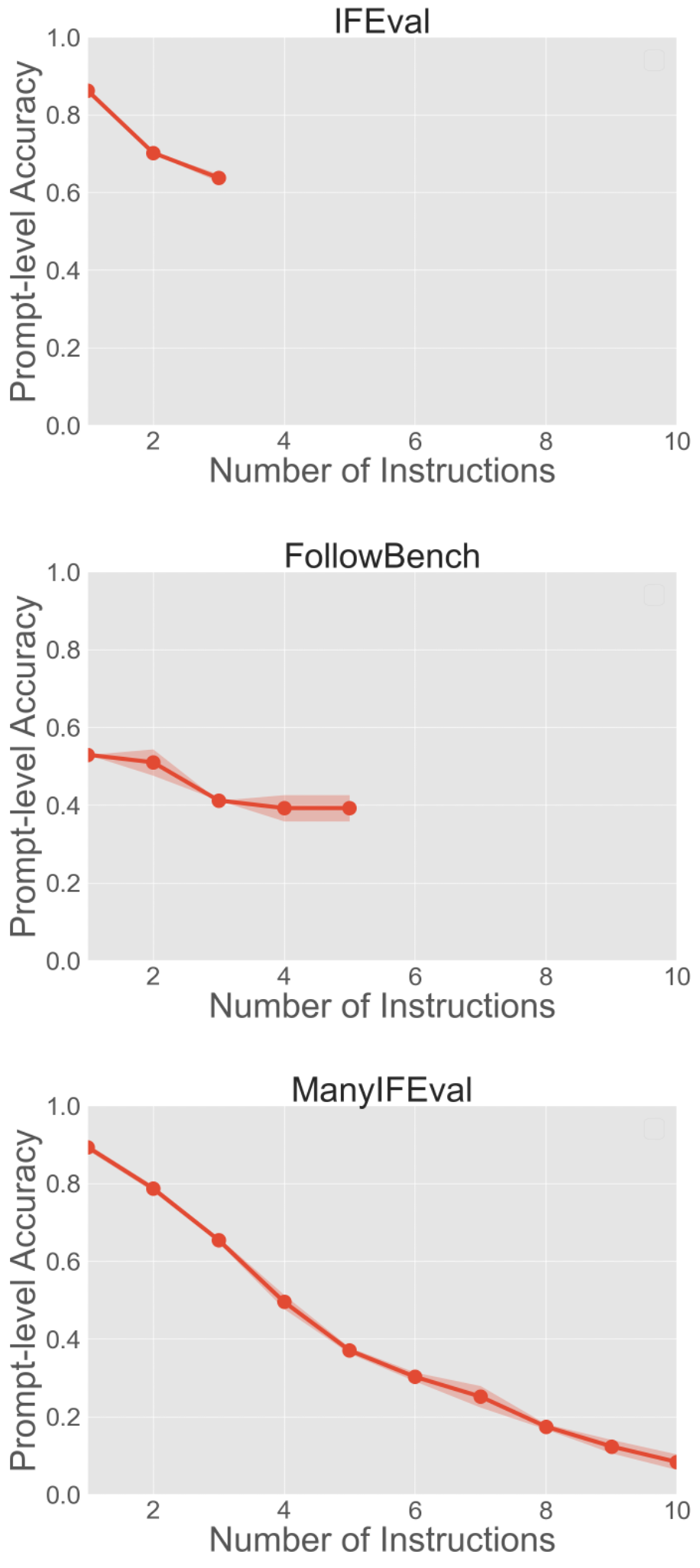}
    \caption{Average Prompt-level accuracy and standard deviation for GPT-4o-mini on IFEval, FollowBench and ManyIFEval across 3 random seeds. The shaded area represents the standard deviation. The results demonstrate ManyIFEval's ability to reliably measure performance trends as instruction counts increase, facilitated by rule-based verification and ample sample sizes.}
    \label{fig:benchmark_result_comparison}
\end{figure}

To illustrate the reliability of ManyIFEval in measuring performance degradation, we evaluated the GPT-4o-mini model. \autoref{fig:benchmark_result_comparison} presents the average Prompt-level Accuracy and standard deviation across three different random seeds on IFEval, FollowBench and ManyIFEval. The consistent trend and minimal standard deviation observed underscore the benchmark's capacity to provide a stable and dependable measure of how an LLM's performance on a given task description is impacted by the addition of multiple, concurrent instructions. 


\section{Details of Evaluation Setup}
\label{app:setup}
We used transformers~\citep{wolf-etal-2020-transformers} and vLLM~\citep{kwon2023efficient} libraries for Gemma 2, Llama 3.1 models' inference. For inference of Qwen2.5-72B, DeepSeek-V3 and DeepSeek-R1, we used endpoints cloud hosted by Fireworks AI~\citep{fireworks}. All models were evaluated using zero-shot prompting presenting the task description along with varying numbers of instructions to each model. During decoding, we employed greedy decoding (top-k=1) to ensure deterministic outputs across models, except for GPT-4o, o3-mini due to API limitations. For open models we tesed locally, we used one H100 GPU(80 GB) and inference for each model takes within 1 hour.

\section{Task Description}
\begin{figure}[t]
    \centering
    \begin{tcolorbox}[title=ManyIFEval Task Descriptions]
    \textbf{Example 1:} Write a blog post about the best way to get a good night's sleep.
    
    \textbf{Example 2:} Are the weather conditions in the Arctic very cold most of the year?
    \end{tcolorbox}
    \caption{Example task descriptions from ManyIFEval benchmark.}
    \label{fig:example_task_description_manyifeval}
\end{figure}

\begin{figure}[t]
    \centering
    \begin{tcolorbox}[title=StyleMBPP Task Descriptions]
    \textbf{Example 1:} Write a python function to find the maximum sum of elements of list in a list of lists. Your code should pass these tests:
    ... (omitted for brevity) ... 
    assert maximum\_Sum([[0,1,3],[1,2,1],[9,8,2],\newline[0,1,0],[6,4,8]]) == 19
    
    \textbf{Example 2:} Write a python function to check whether the first and last characters of a given string are equal or not. Your code should pass these tests:
    ... (omitted for brevity) ...
    assert check\_Equality("mad") == "Not Equal"
    \end{tcolorbox}
    \caption{Example task descriptions from StyleMBPP benchmark.}
    \label{fig:example_task_description_stylembpp}
\end{figure}
\label{app:task_description}
We manually selected 216 task descriptions from the original IFEval~\citep{zhou2023instruction} benchmark for ManyIFEval. Original IFEval's task descriptions are collected by few-shot prompting and manual curation. For StyleMBPP, we used all 500 task descriptions from MBPP benchmark~\citep{mbpp}. MBPP's task descriptions are collected from crowdworkers who have basic knowledge of Python. Details of how the task descriptions are collected can be found in Section 2.1 of MBPP paper~\citep{mbpp}.

We show some example task descriptions in \autoref{fig:example_task_description_manyifeval} and \autoref{fig:example_task_description_stylembpp}. All data are available at \url{https://github.com/kenoharada/Multiple-Instructions-Following}.

\section{Success Rate of Each Instruction Following}
\label{app:each_instruction}
We present the success rates of following each instruction in isolation (when it is the only instruction given alongside the task description) using the GPT-4o model.

\autoref{tab:manyifeval-single} lists the success rates for each type of instruction within ManyIFEval benchmark when presented individually to GPT-4o. Please refer to Table 1 in IFEval paper~\citep{zhou2023instruction} for each instruction's description. We used the same rule-based verifier as \citet{zhou2023instruction}, which are written in Python. Every instruction has its program to judge the instruction is successfully followed or not. We show example rule-based verification code in \autoref{fig:evaluation_code_manyifeval}. \autoref{tab:stylembpp-single} lists the success rates for each type of instruction within StyleMBPP benchmark when presented individually to GPT-4o. \autoref{tab:list-of-verifiable-instruction} shows each instruction's description in StyleMBPP. Similar to ManyIFEval, we used rule-based verifiers written in Python to evaluate whether each instruction is successfully followed or not using Pylint~\citep{pylint}.

Instructions that require inserting specific text (Keywords, Title, MIT License, Quotation, Function docstring) or maintaining a simple pattern (UpperCase, Lowercase, Indentation) are easier for the model to follow. In contrast, instructions related to length or count (e.g., character count, number of sentences, variable name length) tend to have lower success rates.

\begin{table}[h]
\centering
\begin{tabular}{lc}
\hline
\textbf{Instruction} & \textbf{Success Rate} \\
\hline
Keywords to include & 0.98 \\
Keyword occurrence count & 0.96 \\
Forbidden keywords & 0.95 \\
Specific character count & 0.62 \\
Character count & 0.81 \\
Sentence count & 0.73 \\
Paragraph count & 0.87 \\
Placeholder & 0.95 \\
Bullet points & 0.88 \\
Title & 1.0 \\
Uppercase & 0.97 \\
Lowercase & 0.97 \\
Uppercase word count & 0.87 \\
Quotation marks & 1.0 \\
Comma prohibition & 1.0 \\
\hline
\end{tabular}
\caption{Single instruction following success rates for ManyIFEval instructions using GPT-4o.}
\label{tab:manyifeval-single}
\end{table}

\begin{table}[h]
\centering
\begin{tabular}{lc}
\hline
\textbf{Instruction} & \textbf{Success Rate} \\
\hline
MIT License notice & 1.000 \\
Indentation & 1.000 \\
Function docstring & 1.000 \\
Conditional comparison & 0.992 \\
Characters per line & 0.868 \\
Variable name length & 0.794 \\
\hline
\end{tabular}
\caption{Single instruction following success rates for StyleMBPP instructions using GPT-4o.}
\label{tab:stylembpp-single}
\end{table}

\begin{figure*}[t]
    \centering
    \begin{lstlisting}[language=Python, basicstyle=\footnotesize\ttfamily, breaklines=true, frame=single]
class BulletListChecker(Instruction):
  """Checks the bullet list in the prompt."""
  ... (omitted for brevity) ...

  def check_following(self, value):
    """Check if the number of bullet lists meets the requirement.

    Args:
      value: A string representing the response. The response is expected to
        contain some bullet lists that start with `\*`.

    Returns:
      True if the actual number of bullet lists in the response meets the
      requirement.
    """
    bullet_lists = re.findall(r"^\s*\*[^\*].*$", value, flags=re.MULTILINE)
    bullet_lists_2 = re.findall(r"^\s*-.*$", value, flags=re.MULTILINE)
    num_bullet_lists = len(bullet_lists) + len(bullet_lists_2)
    return num_bullet_lists == self._num_bullets
    \end{lstlisting}
    \caption{Python code for evaluating `Bullet points` instruction in ManyIFEval. We used the same code as IFEval~\citep{zhou2023instruction}.}
    \label{fig:evaluation_code_manyifeval}
\end{figure*}

\begin{table*}[t]
\centering
\begin{tabular}{p{4.6cm}p{7.8cm}}
\toprule
\textbf{Instruction Name} & \textbf{Instruction} \\
\midrule
MIT License notice & Ensure the file includes the MIT License notice. \\
\midrule
Indentation & Indent all code blocks using exactly two spaces; do not use tabs. \\
\midrule
Function docstring & Ensure each function has a docstring describing its purpose. \\
\midrule
Conditional comparison & Avoid comparing against True, False, or None using == or !=. Instead, use the variable's truthiness (e.g., `if variable:`) or `is None`/`is not None` for None checks. \\
\midrule
Characters per line & Limit all lines to a maximum of 79 characters. \\
\midrule
Variable name length & All variable names should be at least three characters long. \\
\bottomrule
\end{tabular}
\caption{The list of 6 instructions in StyleMBPP.}
\label{tab:list-of-verifiable-instruction}
\end{table*}

\section{Detailed Experimental and Estimation Results}
\label{app:detailed_results}
In this section, we present detailed aggregate experimental results for Prompt-level Accuracy and Instruction-level Accuracy across all evaluated LLMs on both ManyIFEval and StyleMBPP benchmarks. Additionally, we show detailed results for our performance estimation models when predicting performance on an unseen number of instructions for StyleMBPP.

\subsection{ManyIFEval Evaluation Results}
\begin{figure*}[t]
\begin{center}
\includegraphics[width=\textwidth]{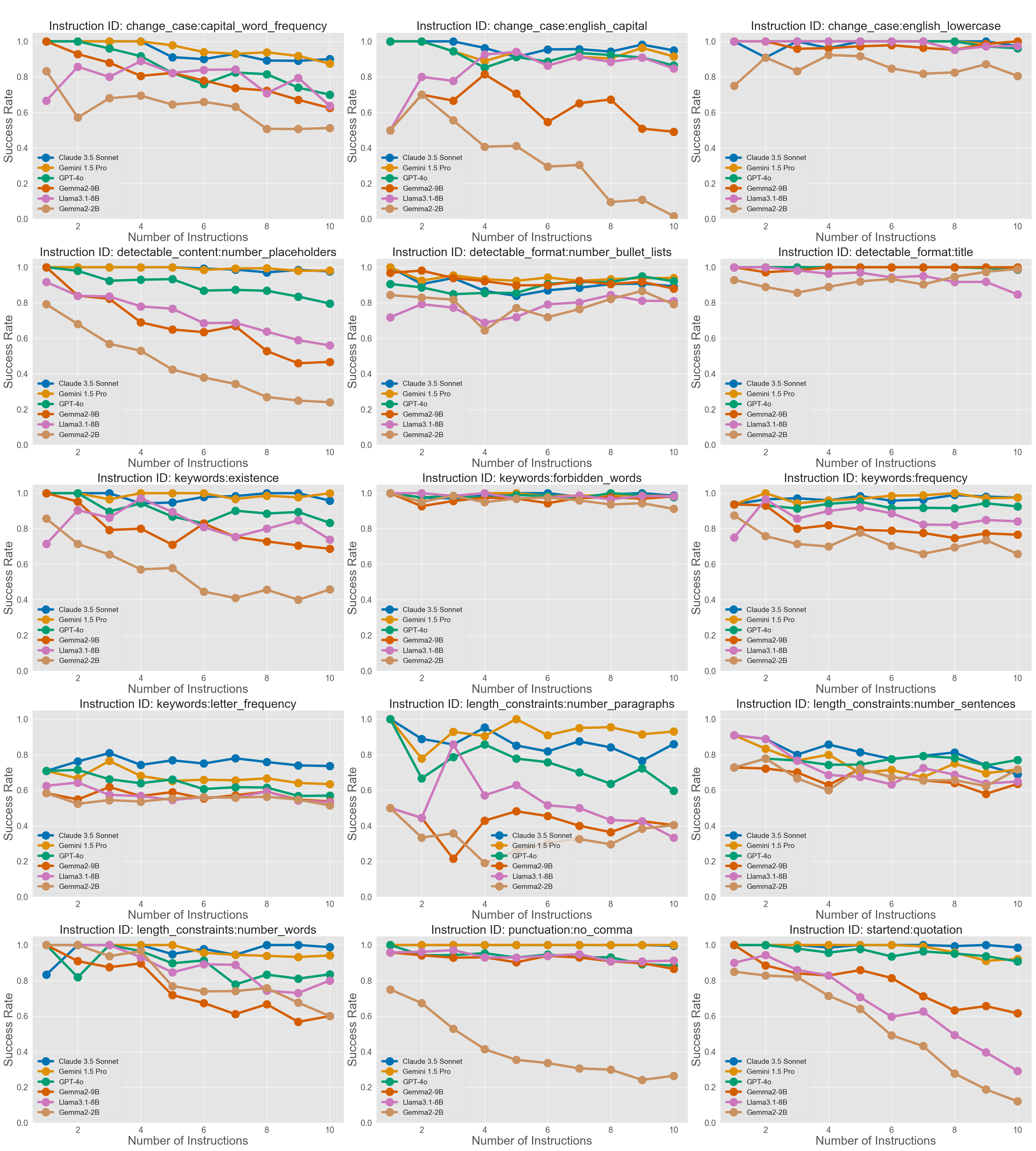}
\end{center}
\caption{\Manyifeval{} instruction-following success rate of each instruction.}
\label{fig:each_instruction_manyifeval}
\end{figure*}

\autoref{tab:manyifeval_prompt_level} and \autoref{tab:manyifeval_instruction_level} provide Prompt-level Accuracy and Instruction-level Accuracy, respectively, for all evaluated LLMs on ManyIFEval benchmark across varying numbers of instructions. \autoref{fig:each_instruction_manyifeval} shows the success rate of following each specific instruction type in ManyIFEval changes as the total number of simultaneously presented instructions increases. We show an example of GPT-4o's response in \autoref{tab:example_zeroshot}.

\begin{table*}
\begin{center}
\begin{tabular}{lllllllllll}
\toprule
Model & n = 1 & n = 2 & n = 3 & n = 4 & n = 5 & n = 6 & n = 7 & n = 8 & n = 9 & n = 10 \\
\midrule
Claude 3.5 Sonnet & 0.95 & 0.90 & 0.88 & 0.79 & 0.72 & 0.68 & 0.67 & 0.62 & 0.58 & 0.48 \\
Gemini 1.5 Pro & 0.96 & 0.89 & 0.86 & 0.78 & 0.71 & 0.63 & 0.58 & 0.57 & 0.45 & 0.39 \\
GPT-4o & 0.94 & 0.85 & 0.73 & 0.64 & 0.57 & 0.46 & 0.39 & 0.39 & 0.29 & 0.21 \\
Gemma2-9B & 0.91 & 0.75 & 0.59 & 0.47 & 0.37 & 0.25 & 0.23 & 0.12 & 0.06 & 0.04 \\
Llama3.1-8B & 0.82 & 0.78 & 0.62 & 0.46 & 0.32 & 0.22 & 0.18 & 0.08 & 0.05 & 0.02 \\
Gemma2-2B & 0.80 & 0.56 & 0.36 & 0.15 & 0.08 & 0.05 & 0.01 & 0.00 & 0.00 & 0.00 \\
Qwen2.5-72B & 0.95 & 0.83 & 0.71 & 0.52 & 0.40 & 0.28 & 0.19 & 0.12 & 0.04 & 0.02 \\
DeepSeek-V3 & 0.96 & 0.90 & 0.85 & 0.70 & 0.63 & 0.48 & 0.43 & 0.33 & 0.29 & 0.19 \\
DeepSeek-R1 & 0.95 & 0.89 & 0.81 & 0.74 & 0.71 & 0.64 & 0.59 & 0.49 & 0.44 & 0.38 \\
o3-mini (low) & 0.97 & 0.93 & 0.92 & 0.84 & 0.81 & 0.75 & 0.69 & 0.70 & 0.64 & 0.53 \\
o3-mini (medium) & 0.98 & 0.95 & 0.94 & 0.89 & 0.87 & 0.83 & 0.78 & 0.73 & 0.73 & 0.64 \\
o3-mini (high) & 1.00 & 0.96 & 0.95 & 0.93 & 0.90 & 0.88 & 0.83 & 0.82 & 0.79 & 0.78 \\
\bottomrule
\end{tabular}
\end{center}
\caption{Prompt-level Accuracy for ManyIFEval}
\label{tab:manyifeval_prompt_level}
\end{table*}

\begin{table*}
\begin{center}
\begin{tabular}{lllllllllll}
\toprule
Model & n = 1 & n = 2 & n = 3 & n = 4 & n = 5 & n = 6 & n = 7 & n = 8 & n = 9 & n = 10 \\
\midrule
Claude 3.5 Sonnet & 0.95 & 0.95 & 0.96 & 0.94 & 0.94 & 0.94 & 0.94 & 0.94 & 0.94 & 0.93 \\
Gemini 1.5 Pro & 0.96 & 0.94 & 0.95 & 0.94 & 0.94 & 0.93 & 0.93 & 0.94 & 0.92 & 0.92 \\
GPT-4o & 0.94 & 0.92 & 0.90 & 0.90 & 0.89 & 0.88 & 0.88 & 0.89 & 0.87 & 0.85 \\
Gemma2-9B & 0.91 & 0.87 & 0.84 & 0.82 & 0.81 & 0.80 & 0.79 & 0.77 & 0.75 & 0.74 \\
Llama3.1-8B & 0.82 & 0.88 & 0.85 & 0.83 & 0.81 & 0.79 & 0.80 & 0.77 & 0.75 & 0.72 \\
Gemma2-2B & 0.80 & 0.75 & 0.71 & 0.65 & 0.65 & 0.62 & 0.60 & 0.58 & 0.57 & 0.55 \\
Qwen2.5-72B & 0.95 & 0.91 & 0.90 & 0.86 & 0.85 & 0.82 & 0.80 & 0.78 & 0.77 & 0.75 \\
DeepSeek-V3 & 0.96 & 0.95 & 0.95 & 0.92 & 0.91 & 0.89 & 0.89 & 0.89 & 0.89 & 0.87 \\
DeepSeek-R1 & 0.95 & 0.94 & 0.94 & 0.92 & 0.93 & 0.92 & 0.93 & 0.92 & 0.92 & 0.92 \\
o3-mini (low) & 0.97 & 0.96 & 0.97 & 0.96 & 0.96 & 0.95 & 0.95 & 0.96 & 0.95 & 0.94 \\
o3-mini (medium) & 0.98 & 0.97 & 0.98 & 0.97 & 0.97 & 0.97 & 0.97 & 0.96 & 0.96 & 0.96 \\
o3-mini (high) & 1.00 & 0.98 & 0.98 & 0.98 & 0.98 & 0.98 & 0.97 & 0.98 & 0.98 & 0.98 \\
\bottomrule
\end{tabular}
\end{center}
\caption{Instruction-level Accuracy for ManyIFEval}
\label{tab:manyifeval_instruction_level}
\end{table*}

\begin{table*}[t]
\centering
\begin{tabular}{p{15truemm}|p{115truemm}}
\hline
\multicolumn{2}{c}{\textbf{Example response of GPT-4o zero-shot for 5 instructions}} \\ 
\hline\hline\\

\textbf{Prompt} & \begin{minipage}{115truemm}
Write a blog post about the best way to get a good night's sleep.\newline

Your response should follow the instructions below:\newline
{\color{red}- Your answer must contain exactly 2 bullet points. Use the markdown bullet points such as: * This is point 1.  * This is point 2}\newline
- In your response, the letter m should appear at least 6 times.\newline
- The response must contain at least 2 placeholders represented by square brackets, such as [address].\newline
- Answer with at least 400 words.\newline
- Do not include keywords ['reason', 'review'] in the response.\newline
\end{minipage}
\\
\hline\\
\textbf{Response} & \begin{minipage}{115truemm} 
\# The Best Way to Get a Good Night's Sleep\newline

Getting a good night's sleep is essential for maintaining overall health and well-being. \newline

... (omitted for brevity) ...\newline

* **Create a Relaxing Bedtime Routine:** Developing a calming pre-sleep routine can signal to your body that it's time to wind down. ... (omitted for brevity) ...\newline

* **Limit Caffeine and Alcohol Intake:** What you consume during the day can significantly impact your sleep quality. ... (omitted for brevity) ...\newline

... (omitted for brevity) ...\newline

* **Invest in a Comfortable Mattress and Pillows:** The quality of your mattress and pillows can make a significant difference in your sleep quality. ... (omitted for brevity) ...\newline

* **Control the Room Temperature:** The temperature of your bedroom can also affect your sleep. ... (omitted for brevity) ...\newline

... (omitted for brevity) ...\newline
For more personalized advice, feel free to reach out to [healthcare professional] or visit [website] for additional resources.\newline
\end{minipage}
\\
\hline
\end{tabular}
\caption{Example response of GPT-4o zero-shot for 5 instructions. Red text indicates instructions that are not being followed. Tested on ManyIFEval.}
\label{tab:example_zeroshot}
\end{table*}

\subsection{StyleMBPP Evaluation Results}
\begin{figure*}[t]
    \begin{center}
    \includegraphics[width=\textwidth]{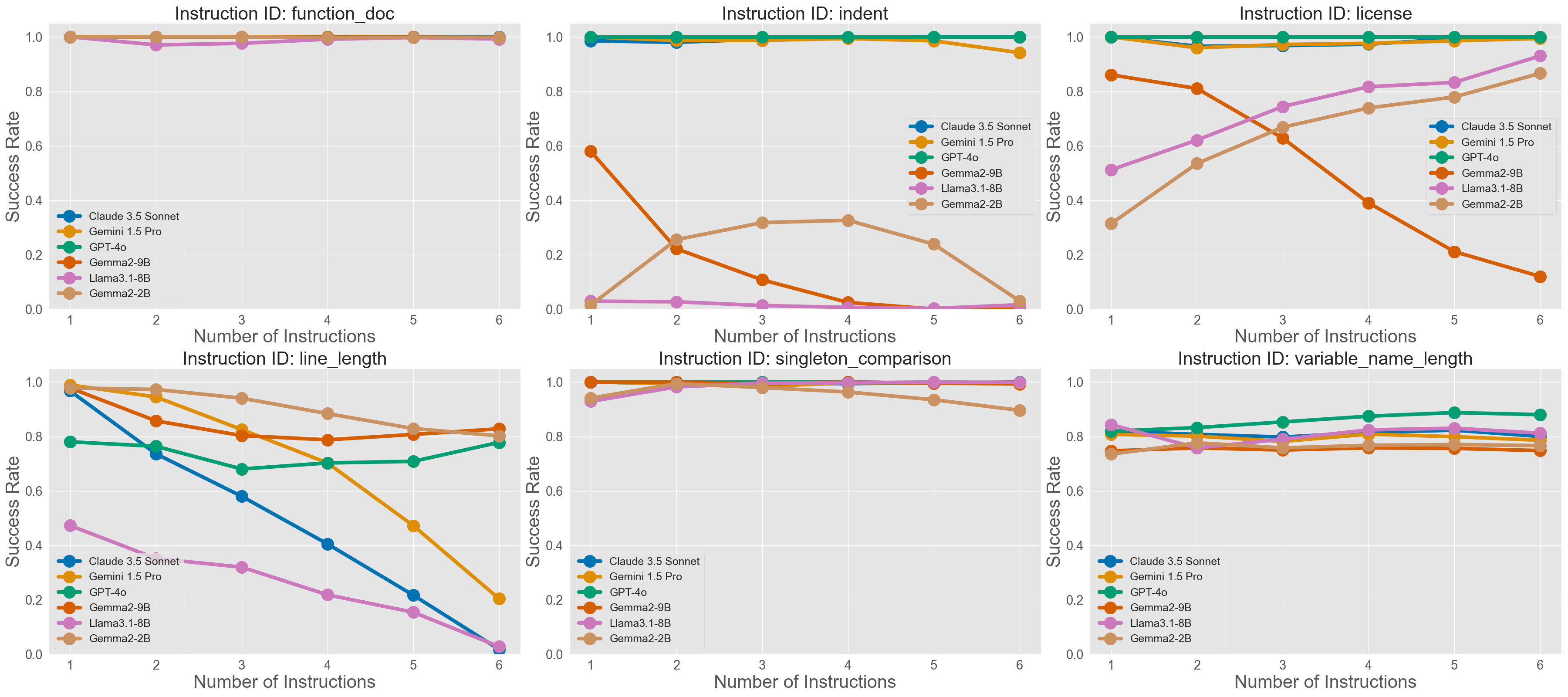}
    \end{center}
    \caption{\Stylembpp{} instruction-following success rate of each instruction.}
    \label{fig:each_instruction_stylembpp}
\end{figure*}

\autoref{tab:stylembpp_prompt_level} and \autoref{tab:stylembpp_instruction_level} provide Prompt-level Accuracy and Instruction-level Accuracy, respectively, for all evaluated LLMs on StyleMBPP benchmark across varying numbers of instructions. \autoref{fig:each_instruction_stylembpp} shows the success rate of following each specific instruction type in StyleMBPP changes as the total number of simultaneously presented instructions increases. 
\subsection{Estimation for Unseen Number of Instructions (StyleMBPP)}
\autoref{tab:stylembpp_unseen_number} shows the performance of logistic regression model in estimating the Prompt-level Accuracy for an unseen number of instructions (n=6) on the StyleMBPP benchmark, based on training with fewer instructions. 

\section{Reasoning Trace Example}
\label{app:reasoning_trace}

\autoref{tab:example_reasoning_trace} demonstrates how reasoning models like DeepSeek-R1 approach multiple-instructions-following tasks. The reasoning trace reveals that the model explicitly identifies and processes each instruction separately, developing a plan before generating the response.

\begin{table*}
\begin{center}
\begin{tabular}{lllllll}
\toprule
Model & n = 1 & n = 2 & n = 3 & n = 4 & n = 5 & n = 6 \\
\midrule
Claude 3.5 Sonnet & 0.96 & 0.84 & 0.67 & 0.48 & 0.29 & 0.01 \\
Gemini 1.5 Pro & 0.97 & 0.90 & 0.79 & 0.68 & 0.45 & 0.13 \\
GPT-4o & 0.93 & 0.86 & 0.76 & 0.72 & 0.68 & 0.68 \\
Gemma2-9B & 0.87 & 0.60 & 0.34 & 0.10 & 0.00 & 0.00 \\
Llama3.1-8B & 0.65 & 0.40 & 0.24 & 0.08 & 0.00 & 0.02 \\
Gemma2-2B & 0.69 & 0.57 & 0.48 & 0.36 & 0.21 & 0.02 \\
Qwen2.5-72B & 0.86 & 0.64 & 0.39 & 0.19 & 0.13 & 0.11 \\
DeepSeek-V3 & 0.92 & 0.77 & 0.56 & 0.32 & 0.18 & 0.01 \\
DeepSeek-R1 & 0.90 & 0.82 & 0.66 & 0.48 & 0.30 & 0.12 \\
o3-mini (low) & 0.96 & 0.86 & 0.68 & 0.56 & 0.34 & 0.18 \\
o3-mini (medium) & 0.95 & 0.84 & 0.69 & 0.57 & 0.38 & 0.15 \\
o3-mini (high) & 0.95 & 0.89 & 0.79 & 0.68 & 0.58 & 0.37 \\
\bottomrule
\end{tabular}
\end{center}
\caption{Prompt-level Accuracy for StyleMBPP}
\label{tab:stylembpp_prompt_level}
\end{table*}

\begin{table*}
\begin{center}
\begin{tabular}{lllllll}
\toprule
Model & n = 1 & n = 2 & n = 3 & n = 4 & n = 5 & n = 6 \\
\midrule
Claude 3.5 Sonnet & 0.96 & 0.91 & 0.88 & 0.86 & 0.84 & 0.80 \\
Gemini 1.5 Pro & 0.97 & 0.95 & 0.92 & 0.91 & 0.87 & 0.82 \\
GPT-4o & 0.93 & 0.93 & 0.92 & 0.93 & 0.93 & 0.94 \\
Gemma2-9B & 0.87 & 0.79 & 0.72 & 0.66 & 0.63 & 0.61 \\
Llama3.1-8B & 0.65 & 0.62 & 0.64 & 0.64 & 0.64 & 0.63 \\
Gemma2-2B & 0.69 & 0.77 & 0.78 & 0.78 & 0.76 & 0.73 \\
Qwen2.5-72B & 0.86 & 0.81 & 0.76 & 0.72 & 0.72 & 0.69 \\
DeepSeek-V3 & 0.92 & 0.89 & 0.84 & 0.80 & 0.77 & 0.74 \\
DeepSeek-R1 & 0.90 & 0.89 & 0.86 & 0.83 & 0.79 & 0.74 \\
o3-mini (low) & 0.96 & 0.93 & 0.89 & 0.88 & 0.85 & 0.82 \\
o3-mini (medium) & 0.95 & 0.92 & 0.89 & 0.89 & 0.85 & 0.82 \\
o3-mini (high) & 0.95 & 0.95 & 0.93 & 0.91 & 0.90 & 0.87 \\
\bottomrule
\end{tabular}
\end{center}
\caption{Instruction-level Accuracy for StyleMBPP}
\label{tab:stylembpp_instruction_level}
\end{table*}

\section{Performance Gap Between Standard and Multiple-Instructions-Following Benchmarks}
\label{app:performance_gap}

To better understand the capability differences between models, we compared their performance on standard benchmarks versus our multiple-instructions-following benchmarks. \autoref{tab:benchmark_performance_comparison} presents results for three representative models across different evaluation tasks. On standard benchmarks (results from~\citet{deepseekv3}) such as MMLU (general knowledge), HumanEval-Mul (coding), and IFEval, the three models demonstrate comparable performance. However, our evaluation on benchmarks reveals substantial performance gap.

\clearpage

\begin{table*}[h]
\centering
\begin{tabular}{lll}
\toprule
Logistic (w/ n) & Abs Err (n=6) & Corr (r)\\
\midrule
trained n$\leq$3& $0.14 \pm 0.22$ & 0.282\\
trained n$\leq$5& $0.09 \pm 0.07$ & 0.924\\
\bottomrule
\end{tabular}
\caption{Result of estimation for unseen number of instructions. Absolute error between predictions and empirical result of successfully following all instructions on unseen number of instructions aggregated over six LLMs on StyleMBPP.}
\label{tab:stylembpp_unseen_number}
\end{table*}

\begin{table*}[t]
\centering
\begin{tabular}{p{15truemm}|p{115truemm}}
\hline
\multicolumn{2}{c}{\textbf{Example reasoning trace of DeepSeek-R1 zero-shot for 5 instructions}} \\ 
\hline\hline\\

\textbf{Prompt} & \begin{minipage}{115truemm}
Write a blog post about the best way to get a good night's sleep.\newline

Your response should follow the instructions below:\newline
- Your answer must contain exactly 2 bullet points. Use the markdown bullet points such as: * This is point 1.  * This is point 2\newline
- In your response, the letter m should appear at least 6 times.\newline
- The response must contain at least 2 placeholders represented by square brackets, such as [address].\newline
- Answer with at least 400 words.\newline
- Do not include keywords ['reason', 'review'] in the response.\newline
\end{minipage}
\\
\hline\\
\textbf{Reasoning trace} & \begin{minipage}{115truemm} 
<think>\newline
Okay, the user wants me to write a blog post about the best way to get a good night's sleep. Let me start by understanding the requirements. They need exactly two bullet points using markdown, the letter 'm' has to appear at least six times, include two placeholders in square brackets, and the total should be over 400 words. Also, I must avoid the words 'reason' and 'review'. \newline
... (omitted for brevity) ...\newline
Let me think of words with 'm'—maybe 'mindfulness', 'environment', 'calming', 'moment', 'maximum', 'minimize'. \newline
... (omitted for brevity) ...\newline
Placeholders: For the first bullet, maybe [specific time] and in the second, [ideal temperature]. \newline
... (omitted for brevity) ...\newline
First bullet: Consistent schedule.\newline
Second bullet: Environment. Mattress, lighting, temperature, mindfulness. \newline
... (omitted for brevity) ...\newline
Okay, that should meet all requirements. Now write it out, keeping paragraphs concise but detailed enough for the word count.\newline
</think>\newline
\end{minipage}
\\
\hline
\end{tabular}
\caption{Example reasoning trace of DeepSeek-R1 zero-shot for 5 instructions. Tested on ManyIFEval.}
\label{tab:example_reasoning_trace}
\end{table*}

\begin{table*}[h]
\centering
\scalebox{0.85}{
\begin{tabular}{lccccc}
\toprule
\multirow{2}{*}{Model} & \multicolumn{3}{c}{Standard Benchmarks} & \multicolumn{2}{c}{Multiple Instructions} \\
\cmidrule(lr){2-4} \cmidrule(lr){5-6}
 & MMLU & HumanEval-Mul & IFEval & ManyIFEval (n=10) & StyleMBPP (n=6) \\
\midrule
DeepSeek-V3 & 88.5 & 82.6 & 86.1 & 0.19 & 0.01 \\
GPT-4o & 87.2 & 80.5 & 84.3 & 0.21 & 0.68 \\
Qwen2.5-72B & 85.3 & 77.3 & 84.1 & 0.02 & 0.11 \\
\bottomrule
\end{tabular}
}
\caption{Performance comparison across standard benchmarks and multiple-instructions-following benchmarks. While models show comparable performance on standard benchmarks (MMLU for general knowledge, HumanEval-Mul for coding, IFEval for single instruction following, results from~\citet{deepseekv3}). However, our evaluation on benchmarks reveals substantial performance gap.}
\label{tab:benchmark_performance_comparison}
\end{table*}

\end{document}

%% file: math_commands.tex
\usepackage{amsmath,amsfonts,bm}

\def\eqref#1{equation~\ref{#1}}

\def\1{\bm{1}}

\DeclareMathAlphabet{\mathsfit}{\encodingdefault}{\sfdefault}{m}{sl}
\SetMathAlphabet{\mathsfit}{bold}{\encodingdefault}{\sfdefault}{bx}{n}

